\documentclass{article}

\usepackage{microtype}
\usepackage{graphicx}
\usepackage{booktabs} 

\usepackage{hyperref}

\usepackage[accepted]{icml2019}

\usepackage[utf8]{inputenc} 
\usepackage[T1]{fontenc}    
\usepackage{url}            
\usepackage{amsfonts}       
\usepackage{nicefrac}       
\usepackage{microtype}      
\usepackage{bm}
\usepackage{hhline}
\usepackage{amsmath,amssymb,amsthm}
\usepackage{enumitem}
\usepackage{subcaption}

\DeclareMathOperator*{\argmin}{argmin}
\DeclareMathOperator*{\argmax}{argmax}

\newtheorem{theorem}{Theorem}

\newtheorem{definition}{Definition}

\newcommand{\ry}[1]{[[\emph{\color{blue}RM: #1}]]}

\newcommand{\norm}[1]{\left\lVert#1\right\rVert}
\newcommand{\eat}[1]{}

\newcommand{\vect}[1]{\bm{#1}}
\newcommand{\matr}[1]{\mathbf{#1}} 
\def\W{{\matr{W}}}
\def\Q{\matr{Q}}
\def\M{\matr{M}}
\def\P{\matr{P}}
\def\II{\matr{I}}
\def\T{\matr{1}}
\def\RR{\matr{R}}

\newcommand{\oracle}{$\text{MARGINAL-ORACLE}$}

\newcommand{\x}{\mathbf{x}}
\newcommand{\q}{\mathbf{q}}
\newcommand{\p}{\mathbf{p}}
\newcommand{\y}{\mathbf{y}}
\newcommand{\z}{\mathbf{z}}

\newcommand{\X}{\mathcal{X}}
\newcommand{\dom}{\mathcal{X}}
\newcommand{\C}{\mathcal{C}}
\newcommand{\marg}{\mathcal{M}}

\newcommand{\bmu}{\boldsymbol{\mu}}

\newcommand{\btheta}{\boldsymbol{\theta}}

\newcommand{\I}{\mathbb{I}}
\newcommand{\R}{\mathbb{R}}

\newcommand{\grad}{\nabla}

\icmltitlerunning{Graphical-model based estimation and inference for differential privacy}

\begin{document}

\twocolumn[
\icmltitle{Graphical-model based estimation and inference for differential privacy}




\begin{icmlauthorlist}
\icmlauthor{Ryan McKenna}{umass}
\icmlauthor{Daniel Sheldon}{umass}
\icmlauthor{Gerome Miklau}{umass}
\end{icmlauthorlist}

\icmlaffiliation{umass}{College of Computer Science, University of Massachusetts, Amherst}

\icmlcorrespondingauthor{Ryan McKenna}{rmckenna@cs.umass.edu}

\icmlkeywords{Differential privacy, graphical models}

\vskip 0.3in
]



\printAffiliationsAndNotice{}  

\begin{abstract}
Many privacy mechanisms reveal high-level information about a data distribution through noisy measurements. It is common to use this information to estimate the answers to new queries. In this work, we provide an approach to solve this estimation problem efficiently using graphical models, which is particularly effective when the distribution is high-dimensional but the measurements are over low-dimensional marginals. We show that our approach is far more efficient than
existing estimation techniques from the privacy literature  and that it can improve the accuracy and scalability of many state-of-the-art mechanisms.
\end{abstract}

\section{Introduction} \label{sec:intro}

Differential privacy \cite{Dwork06Calibrating} has become the dominant standard for controlling the privacy loss incurred by individuals as a result of public data releases.  For complex data analysis tasks, error-optimal algorithms are not known and a poorly designed algorithm may result in much greater error than strictly necessary for privacy.  Thus, careful algorithm design, focused on reducing error, is an area of intense research in the privacy community. 



For the private release of statistical queries, nearly all recent algorithms \cite{zhang2017privbayes,li2015matrix,lee2015maximum,Proserpio12Calibrating,li2014data,qardaji2013understanding,Nikolov13Geometry,hardt2012simple,ding2011differentially,Xiao10Differential,li2010optimizing,hay2010boosting,hardt2010multiplicative,Hardt10Geometry,Barak07Privacy,Gupta11Privately,Thaler12Faster,Acs12Differentially,Zhang14Towards,Yaroslavtsev13Accurate,Cormode12Spatial,Qardaji12Differentially,mckenna2018optimizing} include steps within the algorithm where answers to queries are \emph{inferred} from noisy answers to a set of  \emph{measurement} queries already answered by the algorithm.
 
Inference is a critical component of privacy algorithms because: (i) it can reduce error when answering a query by combining evidence from multiple related measurements, (ii) it provides consistent query answers even when measurements are noisy and inconsistent, and (iii) it provides the above benefits without consuming the privacy-loss budget, since it is performed only on privately-computed measurements without re-using the protected data.  


Consider a U.S. Census dataset, exemplified by the Adult table, which consists of 15 attributes including age, sex, race, income, education.  Given noisy answers to a set of measurement queries,
our goal is to infer answers to one or more new queries.
The measurement queries might be expressed over each individual attribute (age), (sex), (race), etc., as well as selected combinations of attributes (age, income), (age, race, education), etc.  When inference is done properly, the estimate for a new query (e.g., counting the individuals with income>=50K, 10 years of education, and over 40 years old) will use many, or even all, available measurements.


Current inference methods are limited in both scalability and generality.
Most methods first estimate some model of the data and then answer new queries using the model.
Perhaps the simplest model is a full contingency table, which stores a value for every element of the domain. When the measurements are linear queries (a common case, and our primary focus) least-squares \cite{hay2010boosting,Nikolov13Geometry,li2014data,qardaji2013understanding,ding2011differentially,Xiao10Differential,li2010optimizing} and multiplicative-weight updates \cite{hardt2010multiplicative,hardt2012simple} have both been used to estimate this model from the noisy measurements. New queries can then be answered by direct calculation. However, the size of the contingency table is the product of the domain sizes of each attribute, which means these methods break down for high-dimensional cases (or even a modest number of dimensions with large domains).  In the example above, the full contingency table would consist of $10^{19}$ entries.  To avoid this, factored models have been considered \cite{hardt2012simple,zhang2017privbayes}. However, while scalable, these methods have other limitations including restricting the query class \cite{hardt2012simple} or failing to properly account for (possibly varying) noise in measurements \cite{zhang2017privbayes}. 



In this work we show that graphical models provide a foundation for significantly improved inference. We propose to use a graphical model instead of a full contingency table as a model of the data distribution. Doing so avoids an intractable full materialization of the contingency table and retains the ability to answer a broad class of queries. We show that the graphical model representation corresponds to using a maximum entropy criterion to select a single data distribution among all distributions that minimize estimation loss. The structure of the graphical model is determined by the measurements, such that no information is lost relative to a full contingency table representation, but when each measurement is expressible over a low-dimensional marginal of the contingency table, as is common, the graphical model representation is much more compact. 

This work is focused on developing a principled and general approach to inference in privacy algorithms. Our method is agnostic to the loss function used to estimate the data model and to the noise distribution used to achieve privacy.  We focus primarily on linear measurements, but also describe an extension to non-linear measurements


We assume throughout that the measurements are given, but we show our inference technique is versatile since it can be incorporated into many existing private query-answering algorithms that determine measurements in different ways.  For those existing algorithms that scale to high-dimensional data, our graphical-model based estimation method can substantially improve accuracy (with no cost to privacy).  Even more importantly, our estimation method can be added to some algorithms which fail to scale to high-dimensional data, allowing them to run efficiently in new settings.  We therefore believe our inference method can serve as a basic building block in the design of new privacy algorithms.

\section{Background and Problem Statement} \label{sec:background}

\def\alg{{\cal A}}
\def\db{\mathbf{X}}
\def\nbrs{\textrm{nbrs}}

\newcommand{\compl}[1]{{#1}^{-}}

\textbf{Data.}
Our input data represents a population of individuals, each contributing a single record \(\x = (x_1, \ldots, x_d)\) where \(x_i\) is the \(i^{th}\) attribute belonging to a discrete finite domain \(\dom_i\) of \(n_i\) possible values.  The full domain is \(\dom = \prod_{i=1}^d \dom_i\) and its size \(n = \prod_{i=1}^d n_i\) is exponential in the number of attributes. A dataset $\mathbf{X}$ consists of $m$ such records \(\mathbf{X} = (\x^{(1)}, \ldots, \x^{(m)})\). We also consider a normalized contingency table representation $\p$,
which counts the fraction of the population with record equal to \(\x\), for each \(\x\) in the domain. That is, $\p(\x) = \frac{1}{m}\sum_{i=1}^m \I\{\x^{(i)}  = \x\}, \forall \x \in \dom$, where $\I\{ \cdot \}$ is an indicator function. Thus \(\p\) is a probability vector in \(\R^{n}\) with index set \(\dom\) (ordered lexicographically).  We write $\p = \p_{\db}$ when it is important to denote the dependence on $\db$.

\textbf{Queries, Marginals, and Measurements.}
We focus on the most common case of linear queries expressed over subsets of attributes. We will describe an extension to a generalized class of queries, including non-linear ones, in Section~\ref{optimization-in-terms-of-marginals}.
A {\em linear query set} $f_{\Q}(\db)$ is defined by a \emph{query matrix} \(\Q \in \R^{r \times n}\) and has answer \(f_{\Q}(\db) = \Q\, \p_{\db}\). The $i$th row of $\Q$, denoted $\q_i^T$ represents a single scalar-valued query. In most cases we will refer unambiguously to the matrix $\Q$, as opposed to $f_{\Q}$, as the query set. 
We often consider query sets that can be expressed on a {\em marginal} (over a subset of attributes) of the probability vector $\p$.
Let \(A \subseteq [d]\) identify a subset of attributes and, for $\x \in \X$, let \(\x_A = (x_i)_{i \in A}\) be the sub-vector of \(\x\) restricted to \(A\). 
Then the marginal probability vector (or simply ``marginal on A'') \(\bmu_A\), is defined by: 
\[ \bmu_A(\x_A) = \frac{1}{m} \sum_{i=1}^m \I\{ \x^{(i)}_A = \x_A\}, \quad \forall \x_A \in \dom_A := \prod_{i \in A} \dom_i.
\]
The number of entries of the marginal is  $n_A := |\dom_A| = \prod_{i \in A} n_i$, which is exponential in $|A|$ but may be considerably smaller than $n$.
Note that \(\bmu_A(\x_A)\) is a linear function of \(\p\), so there exists a matrix \(\matr{M}_A \in \R^{n_A \times n}\) such that \(\bmu_A = \matr{M}_A\p\).
When a query set depends only on the marginal vector \(\bmu_A\), we call it a \emph{marginal query set} written as \(\Q_A \in \R^{r_A \times n_A}\), and with answer \(f_{\Q_A}(\db) = \Q_A\, \bmu_A\). The marginal query set $\Q_A$ is equivalent to the query set $\Q = \Q_A \matr{M}_A$ on the full contingency table, since $\Q_A \bmu_A = (\Q_A \matr{M}_A) \p$. 
%
%
One marginal query set asks for the marginal vector itself, in which case $\Q_A = \matr{I}_{n_A \times n_A}$ (the identity matrix).

In our problem formulation, we consider measurements consisting of a collection of marginal query sets. Specifically, let \(\C\) be a collection of \emph{measurement sets}, where each $C \in \C$ is a subset of $[d]$.\footnote{Later, these will comprise the cliques of a graphical model, as the notation suggests.}
For each measurement set $C \in \C$, we are given a marginal query set $\Q_C$. 
 The following notation is helpful to refer to combined measurements and their marginals.  Let \(\bmu = (\bmu_C)_{C \in \C}\) be the combined vector of marginals, and let $\Q_{\C}$ be the block-diagonal matrix with diagonal blocks $\{\Q_C\}_{C \in \C}$, so that the entire set of query answers can be expressed as $\Q_{\C} \bmu$. Finally, let $\M_{\C}$ be the matrix that vertically concatenates the matrices $\{\M_C\}_{C \in \C}$, so that $\bmu = \M_{\C} \p$ and $\Q_\C \bmu = \Q_\C \M_\C \p$. This shows that our measurements are equivalent to the combined query set $\Q = \Q_\C \M_\C$ applied to the full table $\p$. 



\textbf{Differential privacy.}
Differential privacy protects individuals by bounding the impact any one individual can have on the output of an admissible  algorithm. This is formalized using the notion of neighboring datasets.
Let $\nbrs(\db)$ denote the set of datasets formed by replacing any $\x^{(i)} \in \db$ with an arbitrary new record $\x'^{(i)} \in \dom$.

\begin{definition}[Differential Privacy;~\citeauthor{Dwork06Calibrating}, \citeyear{Dwork06Calibrating}] \label{def:dp}
A randomized algorithm $\alg$ satisfies $(\epsilon, \delta)$-differential privacy if for any input $\db$, any $\db' \in \nbrs(\db)$, and any subset of outputs $S \subseteq \textrm{Range}(\alg)$, 
$$ \Pr[\alg(\db) \in S] \leq \exp(\epsilon) \Pr[\alg(\db') \in S] + \delta$$
\end{definition}

When $ \delta = 0 $ we say $ \alg $ satisfies $\epsilon$-differential privacy.
Differentially private answers to $f_{\Q}$ are typically obtained with a noise-addition mechanism, such as the Laplace or Gaussian mechanism.
For $\epsilon$-differential privacy, the noise added to the output of $f_{\Q}$ is determined by the {\em $L_1$ sensitivity} of $f_{\Q}$, which, specialized to linear queries, is defined as
$\Delta_{\Q} =  \max_{\db, \db' \in \nbrs(\db)} \| \Q\, \p_{\db} - \Q\, \p_{\db'} \|_1$. It is straightforward to show that $\Delta_{\Q}=\frac{2}{m} \norm{\Q}_1$ where $\norm{\Q}_1$ is the maximum $L_1$ norm of the columns of $\Q$. 
%
%


\begin{definition}[Laplace Mechanism;~\citeauthor{Dwork06Calibrating}, \citeyear{Dwork06Calibrating}] \label{def:laplace}
Given a query set $\Q \in \R^{r \times n}$ of $r$ linear queries, the Laplace mechanism is defined as $\mathcal{L}(\db) = \Q\, \p_{\db} + \mathbf{z}$ where $\mathbf{z} = (z_1, \dots, z_r)$ and each $z_i$ is an i.i.d. random variable from $\text{Laplace}(\Delta_{\Q}/\epsilon)$.
\end{definition}

The Laplace mechanism satisfies $\epsilon$-differential privacy.  The sequential composition property implies that if we answer two query sets $\Q_1$ and $\Q_2$, under $\epsilon_1$ and $\epsilon_2$ differential privacy, respectively, then the combined answers are $(\epsilon_1+\epsilon_2)$-differentially private. The \emph{post-processing} property of differential privacy~\cite{Dwork14Algorithmic} asserts that an algorithm that accepts as input the output of an $\epsilon$-differentially algorithm, but does not use the original protected data, is also $\epsilon$-differentially private.  


\newcommand{\target}{\W} 

\textbf{Problem Statement.}
We assume as given a collection \(\C\) of \emph{measurement sets}, and for each $C \in \C$: a marginal query set $\Q_C$, a privacy parameter $\epsilon_C$, and an $\epsilon_C$-differentially private \emph{measurement} \(\y_C = \Q_C \bmu_C + \text{Lap}(\Delta_{\Q_C}/\epsilon_C)\).  The combined \emph{measurements} are \(\y = (\y_C)_{C \in \C}\) which satisfy $\epsilon$-differential privacy for $\epsilon=\sum_{C\in\C} \epsilon_C$ by sequential composition.
Note that there is no loss of generality in these assumptions; in the extreme case, there may be just a single measurement set $C = [d]$ consisting of all attributes. Formulating the problem this way will allow us to realize computational savings when measurements are not full-dimensional, which is common in practice. We also emphasize that the marginal query set $\Q_C$ is often a complex set of linear queries expressed over measurement set $C$ (not simply a marginal).  Many past works~\cite{li2015matrix,li2014data,qardaji2013understanding,Nikolov13Geometry,ding2011differentially,Xiao10Differential,li2010optimizing,hay2010boosting,Barak07Privacy} have shown that it is beneficial, in the presence of noise-addition for privacy, to measure carefully chosen query sets which balance sensitivity against efficient reconstruction of the workload queries.

Our goal is: given $\y$, derive answers to (possibly different) workload queries $\target$. There are multiple possible motivations: $\target$ may include new queries that were not part of the original measurements; or it is possible that $\target$ is a subset of measurement queries, but we can obtain a more accurate answer by combining \emph{all} of the available information to estimate \(\target \p\) as opposed to just using the noisy answer we got. We describe an extension to non-linear queries and more general linear queries in Section~\ref{optimization-in-terms-of-marginals}; this will be applied to the DualQuery algorithm~\cite{Gaboardi14Dual} in Section~\ref{sec:methods}.

\section{Algorithms for Estimation and Inference}\label{approach}




What principle can we follow to estimate answers to the workload query set? Prior work takes the approach of first using all available information to estimate a full contingency table \(\hat{\p} \approx \p\) and then using \(\hat{\p}\) to answer later queries~\cite{hay2010boosting,li2010optimizing,ding2011differentially,qardaji2013understanding,lee2015maximum}.
We will call finding \(\hat{\p}\) \emph{estimation}, and using \(\hat{\p}\) to answer new queries \emph{inference}.

\subsection{Optimization Formulation} \label{sec:formulation}
The standard framework for estimation and inference is:
\begin{align*}
\hat{\p} &\in \argmin_{\p \in \mathcal{S}} L(\p), & \text{(estimation)} \\
f_{\W}(\db) &\approx \W\, \hat{\p}. &\text{(inference)}
\end{align*}
Here $ \mathcal{S} = \big\{ \p: \p \geq 0, \mathbf{1}^T \p = 1\big\} $  is the probability simplex and $L(\p)$ is a loss function that measures how well $\p$ explains the observed measurements. In past works, $L(\p) = \norm{ \Q \p - \y }$ has been used as a loss function, where $\Q$ is the measured query set and $ \norm{ \cdot } $ is either the $L_1$ norm or $L_2$ norm.  Minimizing the $L_1$ norm is equivalent to maximum likelihood estimation when the noise comes from the Laplace mechanism~\cite{lee2015maximum}.
Minimizing the $L_2$ norm is far more common in the literature however, and it is also the maximum likelihood estimator for Gaussian noise \cite{hay2010boosting,Nikolov13Geometry,li2014data,qardaji2013understanding,ding2011differentially,Xiao10Differential,li2010optimizing, mckenna2018optimizing}. Our method supports both of these loss functions; we only require that $L$ is convex. Both of these loss functions are easily adapted to the situation where queries in $\Q$ may be measured with differing degrees of noise.
The constraint $\p \in \mathcal{S}$ may also be relaxed, which simplifies $L_2$ minimization; additionally, under different assumptions and an alternate version of privacy, the number of individuals may not be known.
%
All existing algorithms to solve these variations of the estimation problem suffer from the same problem: they do not scale to high dimensions since the size of $\p$ is exponential in $d$ and we have to construct it explicitly as an intermediate step even if the inputs and outputs are small (e.g., all measurement queries are over low-dimensional marginals).


\textbf{Optimization in Terms of Marginals.}\label{optimization-in-terms-of-marginals}
For marginal query sets, a loss function will typically depend on $\p$ only through its marginals $\bmu$. For example, when $\Q = \Q_\C \M_\C$ we have $L(\p) = \| \Q\p - \y \| = \| \Q_\C \bmu - \y\| = L(\bmu)$ where we now write the loss function as $L(\bmu)$. More generally, we will consider \emph{any} loss function that only depends on the marginals. A very general case is when $L(\bmu) = -\log p(\y \mid \bmu)$ is the negative log-likelihood of \emph{any} differentially private algorithm that produces output $\y$ that depends only on the marginal vector $\bmu$ (see our treatment of DualQuery\eat{, \citeauthor{Gaboardi14Dual}, \citeyear{Gaboardi14Dual},} in Section~\ref{sec:methods}).

The marginal vector $\bmu$ may be much lower dimensional than $\p$. How can we take advantage of this fact? An ``obvious'' idea would be to modify the optimization to estimate only the marginals as $\hat{\bmu} \in \argmin_{\bmu \in \marg} L(\bmu)$, where \(\marg = \big\{ \bmu: \exists \p \in \mathcal{S} \text{ s.t. } \M_\C \p = \bmu\}\) is the marginal polytope, which is the set of all valid marginals. There are two issues here. First, the marginal polytope has a complex combinatorial structure, and, although it is a convex set, it is generally not possible to enumerate its constraints for use with standard convex optimization algorithms. Note that this optimization problem is in fact a \emph{generic} convex optimization problem over the marginal polytope, and as such it generalizes standard graphical model inference problems~\cite{wainwright2008graphical}. Second, after finding $\hat{\bmu}$ it is not clear how to answer new queries, unless they depend only on some measured marginal \(\bmu_C\).



\textbf{Graphical Model Representation.}
\label{factored-p-and-maximum-entropy-criterion}
After finding an optimal $\hat{\bmu}$ we want to answer new queries that do not necessarily depend directly on the measured marginals.  To do this we need to identify a distribution $\hat{\p}$ that has marginals $\hat{\bmu}$, and we must have tractable representation of this distribution.  Also, since there may be many $\hat{\p}$ that give rise to the same marginals, we want a principled criteria to choose a single estimate, such as the principle of maximum entropy.  We accomplish these goals using undirected graphical models. 

%

\begin{definition}[Graphical model]
  Let \(\p_{\btheta}(\x) = \frac{1}{Z} \exp \big(\sum_{C \in \C} \btheta_C(\x_C) \big)\) be a normalized distribution, where $\btheta_C \in \R^{n_C}$. This distribution is a graphical model that factors over the measurement sets \(\C\), which are the cliques of the graphical model. The vector \(\btheta = (\btheta_C)_{C \in \C}\) is the parameter vector.
  \label{def:graphical-model}
\end{definition}

\begin{theorem}[Maximum entropy \cite{wainwright2008graphical}] \label{theorem:maxent} 
  Given any $\hat{\bmu}$ in the interior of $\marg$ there is a parameter vector \(\hat{\btheta}\) such that the graphical model \(\p_{\hat{\btheta}}(\x)\) has maximum entropy among all $\hat{\p}(\x)$
with marginals \(\hat{\bmu}\).\footnote{If the marginals are on the boundary of $\marg$, e.g., if they contain zeros, there is a sequence of parameters $\{\btheta^{(n)}\}$ such that $\p_{\btheta^{(n)}}(\x)$ converges to the maximum-entropy distribution
    as $n \to \infty$. See~\cite{wainwright2008graphical}.
  }
\end{theorem}
Theorem~\ref{theorem:maxent} says that, after finding $\hat{\bmu}$, we can obtain a factored representation of the maximum-entropy distribution with these marginals by finding the graphical model parameters $\hat{\btheta}$. 
This is the problem of learning in an graphical model, which is well understood~\cite{wainwright2008graphical}.

\subsection{Estimation: optimizing over the marginal polytope}\label{estimation}

\begin{algorithm}[tb]
    \caption{Proximal Estimation Algorithm} \label{alg:proximal2}
\begin{algorithmic}
    \STATE {\bfseries Input:} Loss function $L(\bmu)$ between $\bmu$ and $\y$
    \STATE {\bfseries Output:} Estimated data distribution $\hat{\p}_{\btheta}$
    \STATE $\btheta = \vect{0}$
    \FOR{$t=1, \dots, T$}
        \STATE $\bmu = \oracle(\btheta)$
        \STATE $\btheta = \btheta - \eta_t \grad L(\bmu)$
    \ENDFOR
    \STATE {\bfseries return} $\hat{\p}_{\btheta}$
\end{algorithmic}
\end{algorithm}

We need algorithms to find \(\hat{\bmu}\) and \(\hat{\btheta}\).  We considered a variety of algorithms and present two of them here.  Both are proximal algorithms for solving convex problems with ``simple'' constraints \cite{parikh2014proximal}.  Central to our algorithms is a subroutine \oracle{}, which is some black-box algorithm for computing the clique marginals $\bmu$ of a graphical model from the parameters $\btheta$.  This is the problem of \emph{marginal inference} in a graphical model.  \oracle{} may be any marginal inference routine --- we use belief propagation on a junction tree. In the remainder of this section, we assume that the clique set $\C$ are the cliques of a junction tree. This is without loss of generality, since we can enlarge cliques as needed until this property is satisfied.

Algorithm~\ref{alg:proximal2} is a routine to find $\hat{\bmu}$ by solving a convex optimization problem over the marginal polytope. Due to the special structure of the algorithm it also finds the parameters $\hat{\btheta}$. Algorithm~\ref{alg:proximal2} is inspired by the entropic mirror descent algorithm for solving convex optimization problems over the probability simplex \cite{beck2003mirror}.  The iterates of the optimization are obtained by solving simpler optimization problems of the form: \vspace{-1ex}
\begin{equation} \label{eq:update}
\bmu^{t+1} = \argmin_{\bmu \in \marg} \bmu^T \grad L(\bmu^t) + \frac{1}{\eta_t} D(\bmu, \bmu^t)
\end{equation}
where $D$ is a Bregman divergence that is chosen to reflect the geometry of the marginal polytope.  Here we use the following Bregman divergence generated from the Shannon entropy: $ D(\bmu, \bmu^t) = -H(\bmu) + H(\bmu^t) + (\bmu - \bmu^t)^T \grad H(\bmu^t) $, where $H(\bmu)$ is the Shannon entropy of the graphical model $\p_{\btheta}$ with marginals $\bmu$. Since we assumed above that $\bmu$ are marginals of the cliques of a junction tree, the Shannon entropy is convex and easily computed  as a function of $\bmu$ alone~\cite{wainwright2008graphical}.\footnote{An alternative would be to use the Bethe entropy as in ~\cite{vilnis2015bethe}. The Bethe entropy is convex and computable from $\bmu$ alone regardless of the model structure. Using Bethe entropy would lead to approximate marginal inference instead of exact marginal inference as the subproblems, which is an interesting direction for future work.}

%
%
With this divergence, the objective of the subproblem in Equation \ref{eq:update} can be seen to be equal to a variational free energy, which is minimized by marginal inference in a graphical model.
The full derivation is provided in the supplement.  The implementation of Algorithm \ref{alg:proximal2} is very simple --- it simply requires calling \oracle{} at each iteration.  Additionally, even though the algorithm is designed to find the optimal $\bmu$, it also returns the corresponding graphical model parameters $\btheta$ ``for free'' as a by-product of the optimization.  This is evident from Algorithm \ref{alg:proximal2}: upon convergence, $\bmu$ is the vector of marginals of the graphical model with parameters $\btheta$. The variable $\eta_t$ in this algorithm is a step size, which can be constant, decreasing, or found via line search.  This algorithm is an instance of mirror descent, and thus inherits its convergence guarantees.  It will converge for any convex loss function $L$ at a $O(1 / \sqrt{t})$ rate,\footnote{That is, $L(\bmu^t) - L(\bmu^*) \in O(1 / \sqrt{t})$.} even ones that are not smooth, such as the $L_1$ loss.

We now present a related algorithm which is based on the same principles as Algorithm \ref{alg:proximal2} but has an improved $O(1/t^2)$ convergence rate for convex loss functions with Lipchitz continuous gradients.  Algorithm~\ref{alg:proximal} is based on Nesterov's accelerated dual averaging approach \cite{nesterov2009primal,xiao2010dual,vilnis2015bethe}.  The per-iteration complexity is the same as Algorithm \ref{alg:proximal2} as it requires calling the \oracle{} once, but this algorithm will converge in fewer iterations.  Algorithm \ref{alg:proximal} has the advantage of not requiring a step size to be set, but it requires knowledge of the Lipchitz constant of $ \grad L$.  For the standard $L_2$ loss with linear measurements, this is equal to the largest eigenvalue of $ \Q^T \Q $.  The derivation of this algorithm appears in the supplement.

\begin{algorithm}[tb]
    \caption{Accelerated Proximal Estimation Algorithm} \label{alg:proximal}
\begin{algorithmic}
    \STATE {\bfseries Input:} Loss function $L(\bmu)$ between $\bmu$ and $\y$
    \STATE {\bfseries Output:} Estimated data distribution $\hat{\p}_{\btheta}$
    \STATE $K = $ Lipchitz constant of $\grad L$
    \STATE $\bar{\vect{g}} = \vect{0}$
    \STATE $\vect{\nu}, \bmu = \oracle(\vect{0})$
    \FOR{$t=1, \dots, T$}
        \STATE $c = \frac{2}{t+1}$
        \STATE $\vect{\omega} = (1-c) \bmu + c \vect{\nu}$
        \STATE $\bar{\vect{g}} = (1-c) \bar{\vect{g}} + c \grad L(\vect{\omega})$
        \STATE $\btheta = \frac{-t (t+1)}{4 K} \bar{\vect{g}} $
        \STATE $\vect{\nu} = \oracle(\btheta)$
        \STATE $\bmu = (1-c) \bmu + c \vect{\nu}$
    \ENDFOR
    \STATE {\bfseries return} graphical model $\hat{\p}_{\btheta}$ with marginals $\bmu$
\end{algorithmic}
\end{algorithm}

\subsection{Inference}\label{inference}

Once $\hat{\p}_{\btheta}$ has been estimated, we need algorithms to answer new queries without materializing the full contingency table representation.  This corresponds to the problem of inference in a graphical model. If the new queries only depend on $\hat{\p}_{\btheta}$ through its clique marginals $\bmu$, we can immediately answer them using \oracle{}, or by saving the final value of $\bmu$ from Algorithms~\ref{alg:proximal2} or \ref{alg:proximal}. If the new queries depend on some other marginals outside of the cliques of the graphical model, we instead use the variable elimination algorithm \cite{koller2009probabilistic} to first compute the necessary marginal, and then answer the query.  In Section \ref{sec:inference} of the supplement, we present a novel inference algorithm that is related to variable elimination but is faster for answering certain queries because it does not need to materialize full marginals if the query does not need them.  For more complicated downstream tasks, we can generate synthetic data by sampling from $\hat{\p}_{\btheta}$, although this should be avoided when possible as it introduces additional sampling error.

\section{Use in Privacy Mechanisms} \label{sec:methods}

Next we describe how our estimation algorithms can improve the accuracy and/or scalability of four state-of-the-art mechanisms: MWEM, PrivBayes, HDMM, and DualQuery.

\textbf{MWEM.} The multiplicative weights exponential mechanism \cite{hardt2012simple} is an active-learning style algorithm that is designed to answer a workload of linear queries.  MWEM maintains an approximation of the data distribution and at each time step selects the worst approximated query $\q_i^T$ from the workload via the exponential mechanism \cite{mcsherry2007mechanism}. It then measures the query using the Laplace mechanism as $y_i = \q_i^T \p + z_i$ and then updates the approximate data distribution by incorporating the measured information using the multiplicative weights update rule.
The most basic version of MWEM represents the approximate data distribution in vector form, and updates it according to the following formula after each iteration:
%
\begin{equation} \label{eq:mw}
\hat{\p} \leftarrow \hat{\p} \odot \exp{(-\q_i (\q_i^T \hat{\p} - y_i) / 2m)} / Z,
\end{equation}
where $\odot$ is elementwise multiplication and $Z$ is a normalization constant.

It is infeasible to represent $\p$ explicitly for high-dimensional data, so this version of MWEM is only applicable to relatively low-dimensional data.  Hardt et al describe an enhanced version of MWEM, which we call \emph{factored MWEM}, that is able to avoid materializing this vector explicitly, in the special case when the measured queries decompose
over disjoint subsets of attributes.  
In that case, $\p$ is represented implicitly as a product of independent distributions over smaller domains, i.e., $\p(\x) = \prod_{C \in \C} \p_C(\x_C) $, and the update is done on one group at a time.
However, this enhancement breaks down for measurements on overlapping subsets of attributes in high-dimensional data, so MWEM is still generally infeasible to run except on simple workloads.

We can replace the multiplicative weights update with a call to Algorithm \ref{alg:proximal} using the standard $L_2$ loss function (on all measurements up to that point in the algorithm).  By doing so, we learn a compact graphical model representation of $\hat{\p}$, which avoids materializing the full $\p$ vector even when the measured queries overlap in complicated ways.  This allows MWEM to scale better and run in settings where it was previously infeasible. We remark that Equation \ref{eq:mw} is closely related to the update equation for entropic mirror descent \cite{beck2003mirror}, suggesting that if the update equation is iterated until convergence, it solves the same $L_2$ minimization problem that we consider.  More details on this are given in Section ~\ref{sec:mw_emd} of the supplement.

\textbf{PrivBayes.} PrivBayes \cite{zhang2017privbayes} is a differentially private mechanism that generates synthetic data.  It first spends half the privacy budget to learn a Bayesian network structure that captures the dependencies in the data, and then uses the remaining privacy budget to measure the statistics---which are marginals---necessary to learn the Bayesian network parameters.  PrivBayes uses a heuristic of truncating negative entries of noisy measurements and normalizing to get conditional probability tables. It then samples a synthetic dataset of $m$ records from the Bayesian network from which consistent answers to workload queries can be derived.  While this is simple and efficient, the heuristic does not properly account for measurement noise and sampling may introduce unnecessary error.

We can replace the PrivBayes estimation and sampling step with a call to Algorithm ~\ref{alg:proximal}, using an appropriate loss function (e.g. $L_1$ or $L_2$), to estimate a graphical model. 
Then we can answer new queries by performing graphical model inference (Section~\ref{inference}), rather than using synthetic data. 



\textbf{HDMM. } The high-dimensional matrix mechanism \cite{mckenna2018optimizing} is designed to answer a workload of linear queries on multi-dimensional data. It selects the set of measurements that minimizes estimated error on the input workload.  The measurements are then answered using the Laplace mechanism, and inconsistencies resolved by solving an ordinary least squares problem of the form:
$ \hat{\p} = \argmin \norm{ \Q \p - \y }_2 $.
Solving this least squares problem is the main bottleneck of HDMM, as it requires materializing the data vector even when $\Q$ contains queries over the marginals of $\p$.

We can replace the HDMM estimation procedure with  Algorithm~\ref{alg:proximal}, using the same $L_2$ loss function.  If the workload contains queries over low-dimensional marginals of $\p$, then $\Q$ will contain measurements over the low-dimensional marginals too.
Thus, we replace the full ``probability'' vector $\hat{\p}$ with a graphical model $\hat{\p}_{\btheta}$.
Also $\hat{\p}$ may contain negative values and need not sum to $1$ since HDMM solves an \emph{ordinary} (unconstrained) least squares problem.

\textbf{DualQuery. } DualQuery \cite{Gaboardi14Dual} is an iterative algorithm inspired by the same two-player game underlying MWEM. It generates synthetic data to approximate the true data on a workload of linear queries.  DualQuery maintains a distribution over the workload queries that depends on the true data so that poorly approximated queries have higher probability mass.  In each iteration, samples are drawn from the query distribution, which are proven to be differentially private.  The sampled queries are then used to find a single record from the data domain (without accessing the protected data), which is added to the synthetic database.

The measurements --- i.e., the random outcomes from the privacy mechanism --- are the queries sampled in each iteration.  Even though these are very different from the linear measurements we have primarily focused on, we can still express the log-likelihood as a function of $\p$ and select $\p$ to maximize the log-likelihood using Algorithm ~\ref{alg:proximal2} or \ref{alg:proximal}. The log-likelihood only depends on $\p$ through the answers to the workload queries. If the workload can be expressed in terms of $\bmu$ instead, the log-likelihood can as well.
Thus, after running DualQuery, we can call Algorithm \ref{alg:proximal2} with this custom loss function to estimate the data distribution, which we can use in place of the synthetic data produced by DualQuery.  The full details are given in the supplementary material.

\section{Experimental evaluation} \label{sec:experiments}

In this section, we measure the accuracy and scalability improvements enabled by probabilistic graphical-model (PGM) based estimation when it is incorporated into existing privacy mechanisms.  

\subsection{Adding PGM estimation to existing algorithms}

We run four algorithms: MWEM, PrivBayes, HDMM, and DualQuery, with and without our graphical model technology using a privacy budget of $ \epsilon = 1.0 $ (and $\delta = 0.001$ for DualQuery).  We run Algorithm \ref{alg:proximal2} with line search for DualQuery and Algorithm \ref{alg:proximal} for the other mechanisms, each for 10000 iterations.  We repeat each experiment five times and report the median workload error. Experiments are done on 2 cores of a single compute cluster node with 16 GB of RAM and 2.4 GHz processors.

We use a collection of four multi-dimensional datasets in our experiments, summarized in Table \ref{table:datasets}.  Each dataset consists of a collection of categorical and numerical attributes (with the latter discretized into 100 bins).  Note the large domain of each dataset, which is the main property that makes efficient estimation challenging.  

\begin{table}[]
\caption{Datasets used in experiments along with the number of queries in the workload used with the dataset.} \label{table:datasets}
\begin{tabular}{c|cccc}
\textbf{Dataset} & \textbf{Records} & \textbf{Attributes} & \textbf{Domain} & \textbf{Queries} \\\hline
\textbf{Titanic} & 1304             & 9                   & 3e8  & 4851                \\
\textbf{Adult}   & 48842            & 15                  & 1e19 & 62876                \\
\textbf{Loans}   & 42535            & 48                  & 5e80 & 362201                \\
\textbf{Stroke}  & 19434           & 110                 & 4e104 & 17716               \\
\end{tabular}
\end{table}

\begin{figure*}[t]
  \centering
  \begin{subfigure}[b]{13.1em}
    \centering\includegraphics[width=13.15em]{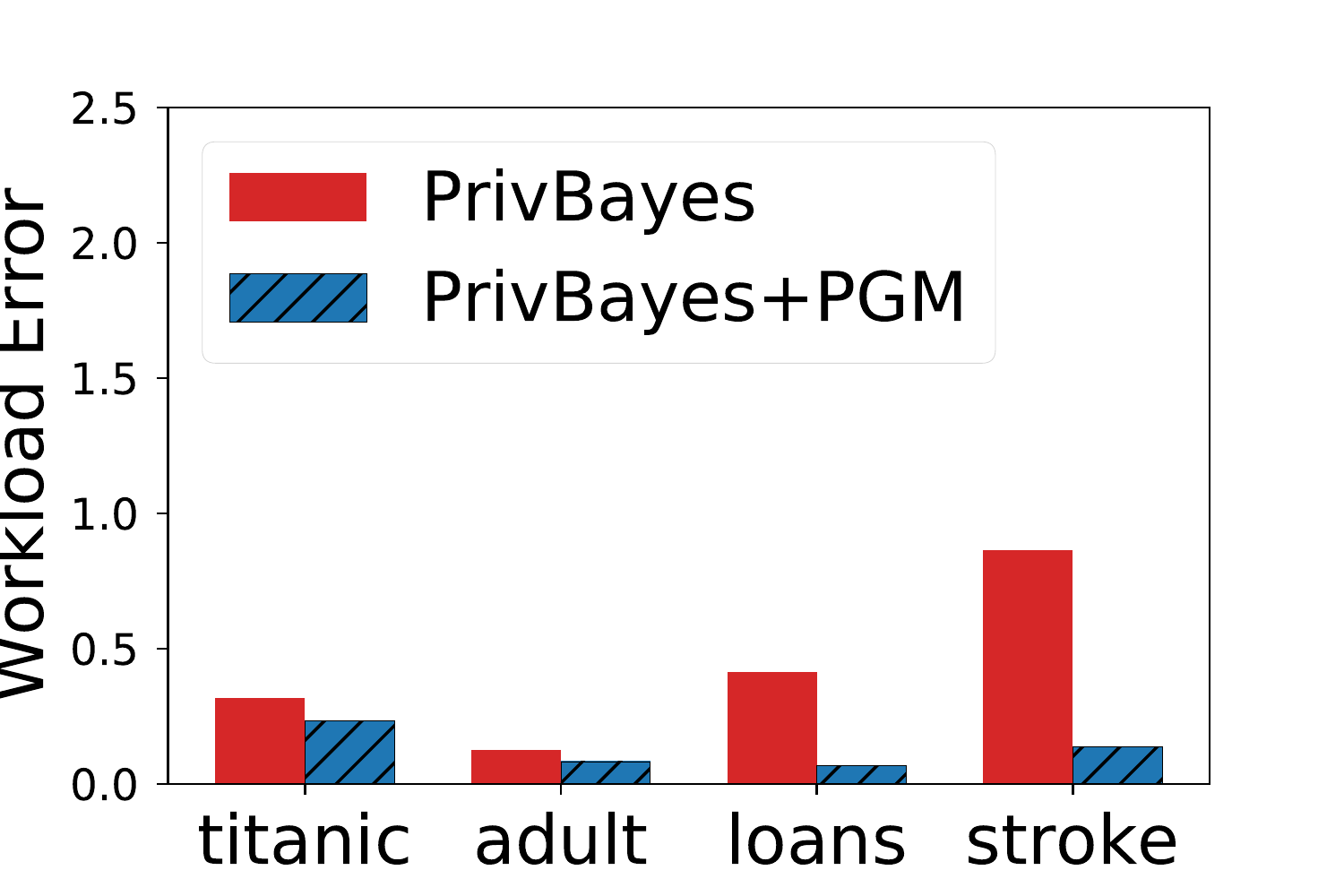} 
    \caption{\label{fig:privbayes} PrivBayes}
  \end{subfigure}%
  \hspace{-1.5em}
  \begin{subfigure}[b]{13.1em}
    \centering\includegraphics[width=13.15em]{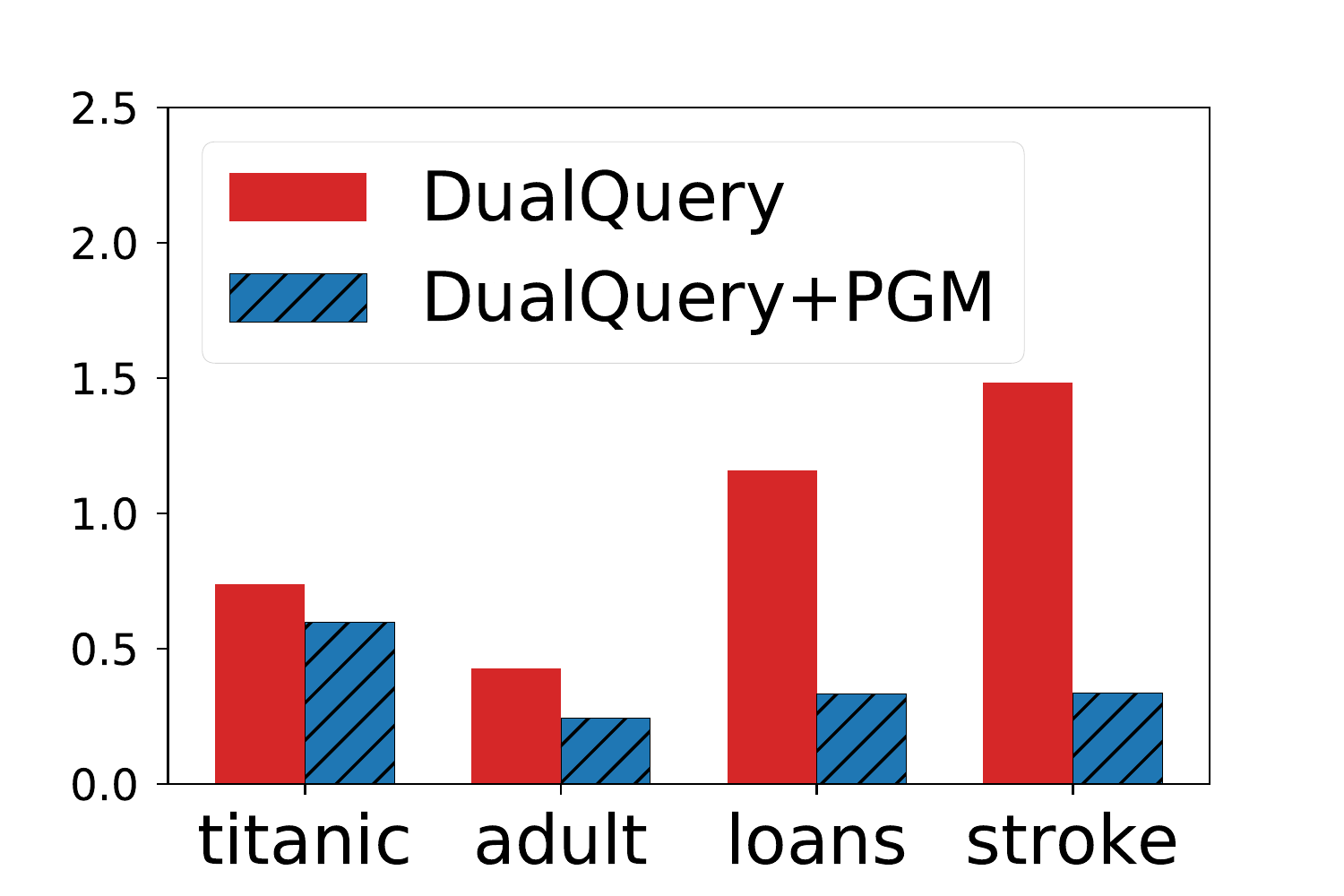}
    \caption{\label{fig:dq} DualQuery}
  \end{subfigure}%
  \hspace{-1.5em}
  \begin{subfigure}[b]{13.1em}
    \centering\includegraphics[width=13.15em]{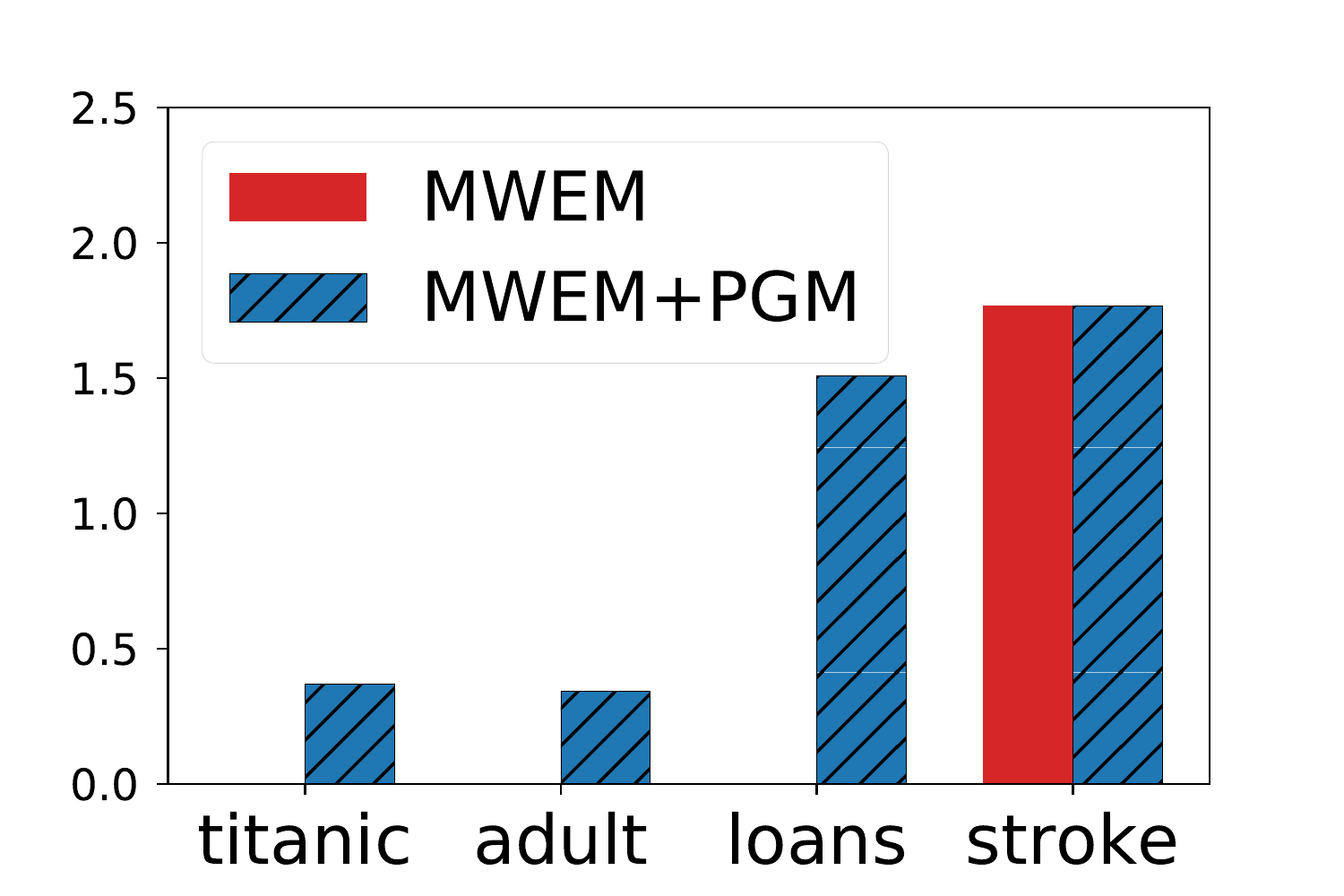} 
    \caption{\label{fig:mwem} MWEM}
  \end{subfigure}%
  \hspace{-1.5em}
  \begin{subfigure}[b]{13.1em}
    \centering\includegraphics[width=13.15em]{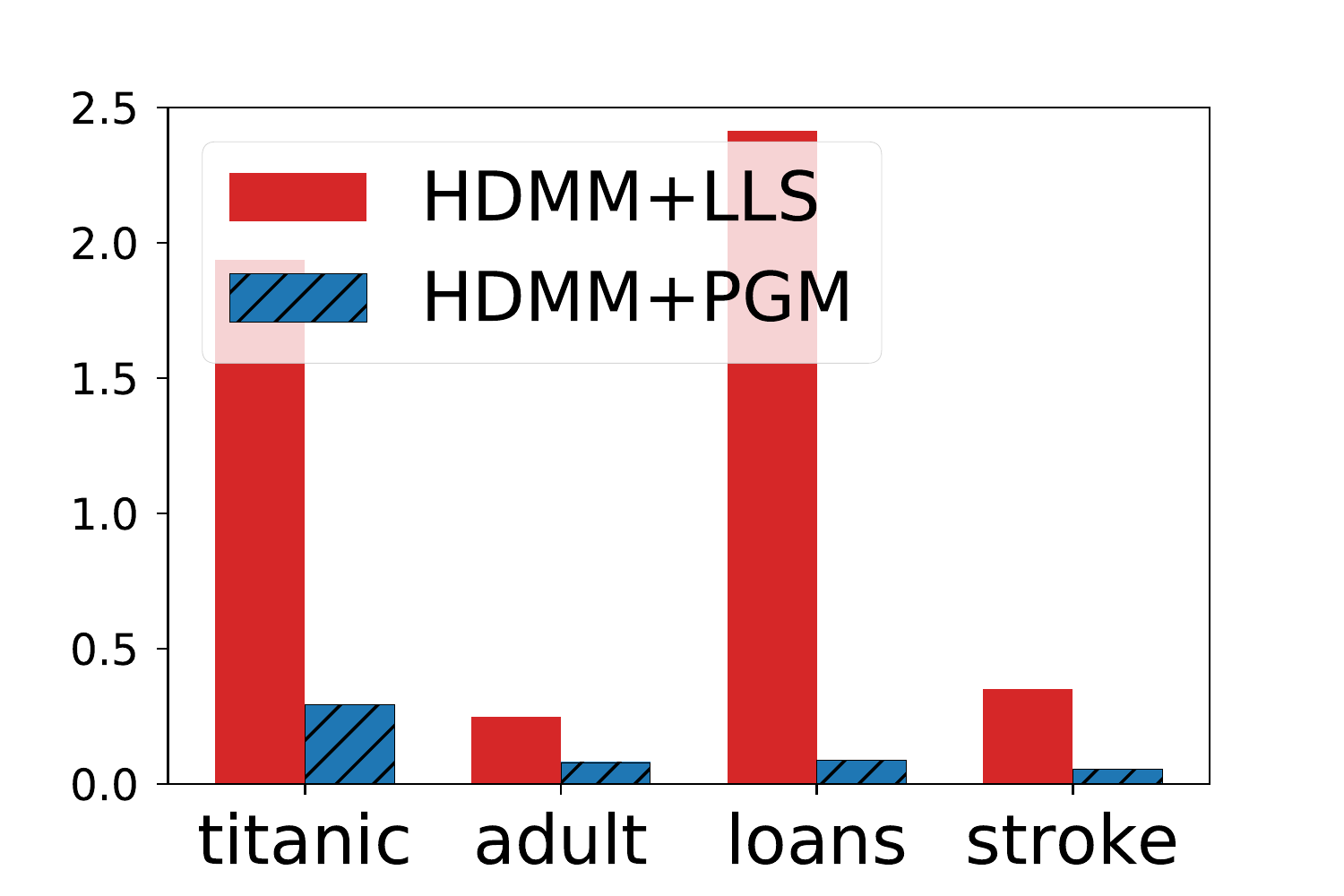} 
    \caption{\label{fig:hdmm} HDMM}
  \end{subfigure}%
 \caption{Workload error of four mechanisms on four datasets, with and without our PGM estimation algorithm for $\epsilon = 1.0$.} \label{fig:accuracy}
\end{figure*}


For each dataset, we construct a workload of counting queries which is an extension of the set of three-way marginals. First, we randomly choose 15 subsets of attributes of size 3, $\C$. For each subset $C \in \C$, if $C$ contains only categorical attributes, we define sub-workload $\W_C$ to be a 3-way marginal.  However,  when $C$ contains any discretized numerical attributes, we replace the set of unit queries used in a marginal with the set of prefix range queries.  For example, if $C=\langle$sex, education, income$\rangle$ then the resulting subworkload $\W_C$ would consist of all queries of the form:
$\mbox{sex}=x, \mbox{education}=y, \mbox{income} \in [0,z]$ where $x,y,z$ range over the domains of the attributes, respectively.
The final workload is the union of the 15 three-way subworkloads defined above.

We measure the error on the workload queries as:
$$ Error = \frac{1}{| \C |} \sum_{C \in \C} \frac{ \norm{ \W_{C} \bmu_C - \W_{C} \hat{\bmu}_C }_1}{2 \norm{ \W_C \bmu_C }_1} $$
where the summand is related to the total variation distance (and is equal in the special case when $\W_C = \II$).  

\textbf{Improved accuracy.}
PrivBayes and DualQuery are highly scalable algorithms supporting the large domains considered here.  Figures ~\ref{fig:privbayes} and \ref{fig:dq} show that incorporating PGM estimation significantly improves accuracy.
For PrivBayes, workload error is reduced by a factor of 6$\times$ and 7$\times$ on the Loans and Stroke datasets, respectively, and a modest $30\%$ for Adult. For DualQuery, we also observe very significant error reductions of 1.2$\times$, 1.8$\times$, 3.5$\times$, and 4.4$\times$.

\textbf{Replacing infeasible estimation methods.}
The MWEM and HDMM algorithms fail to run on the datasets and workloads we consider because both require representations too large to maintain in memory.  However, incorporating PGM estimation makes these algorithms feasible. 

As Figure \ref{fig:mwem} shows, for the first three datasets, MWEM crashed before completing because it ran out of memory or timed out.  For example, on one run of the Adult dataset, the first three chosen queries were on the (race, native-country, income), (workclass, race, capital-gain), and (marital status, relationship, capital-gain) marginals.  Since these all overlap with respect to race and capital-gain, factored MW offers no benefit and the entire vector $\p_{C}$ must be materialized over these attributes, which requires over 100 MB. After 5 iterations, the representation requires more than 2 GB, at which point it timed out.  Interestingly, MWEM was able to run on the stroke dataset, which has the largest domain and greatest number of attributes.  This is mainly because the workload did not contain as many queries involving common attributes.  In general, MWEM's representation will not explode as long as the workload (and therefore its measurements) consist solely of queries defined over low-dimensional marginals that do not have common attributes. Unfortunately this imposes a serious restriction on the workloads MWEM can support. 


Although the HDMM algorithm fails to run, for the purpose of comparison, we run a modified version of the algorithm (denoted HDMM+LLS) which uses local least squares independently over each measurement set instead of global least squares over the full data vector.  While scalable, Figure~\ref{fig:hdmm} shows that this estimation is substantially worse than PGM estimation, especially on the titanic and loans dataset.  Incorporating PGM estimation offers error reductions of 6.6$\times$, 3.2$\times$, 27$\times$, and 6.3$\times$ on the four datasets.  These improvements primarily stem from non-negativity and global consistency. 

\begin{figure}[t]
\centering
\begin{subfigure}[b]{11.7em}
    \centering\includegraphics[width=11.7em]{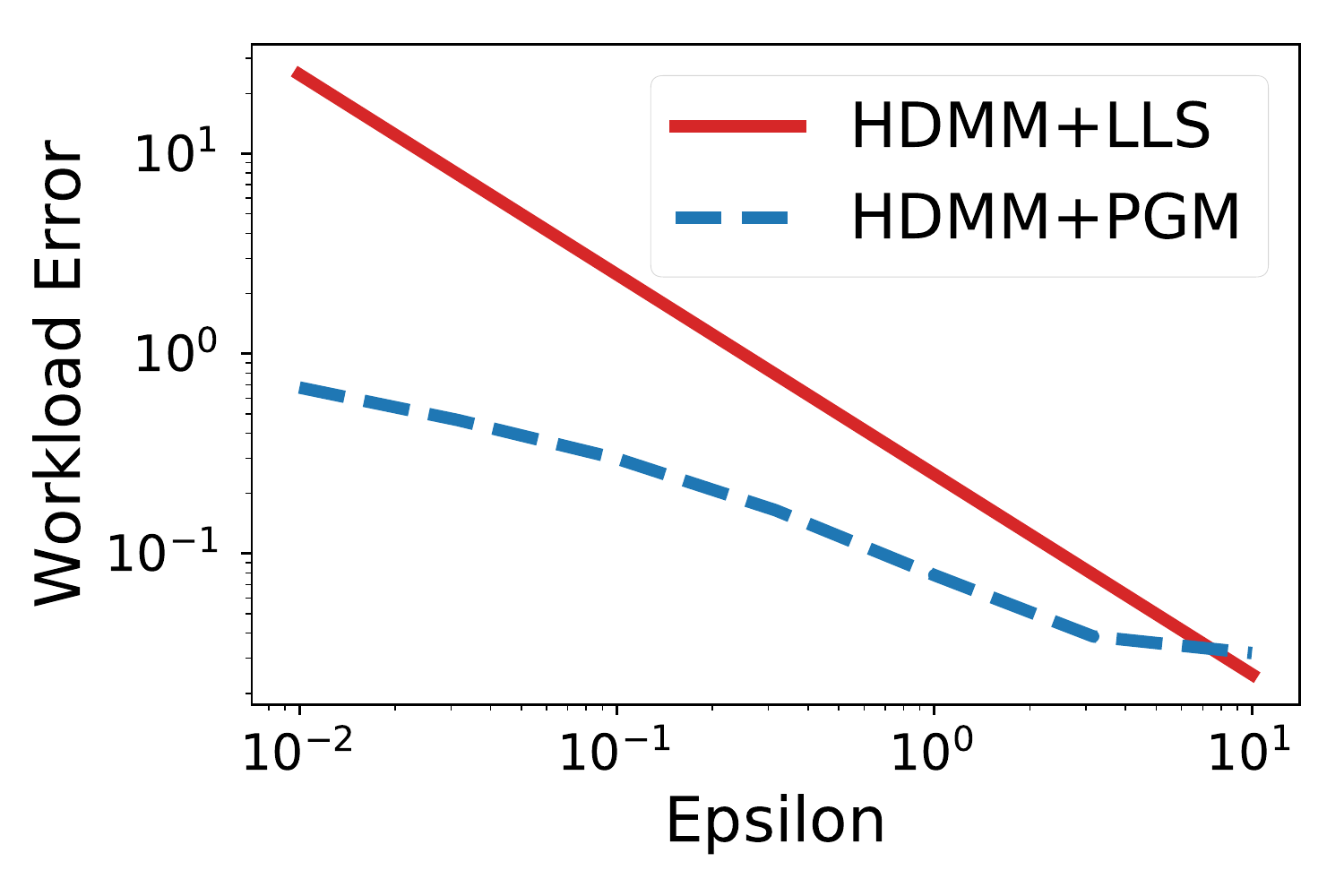} \vspace{-1.5em}
    \caption{\label{fig:sensitivity}}
\end{subfigure}%
\begin{subfigure}[b]{11.7em}
    \centering\includegraphics[width=11.7em]{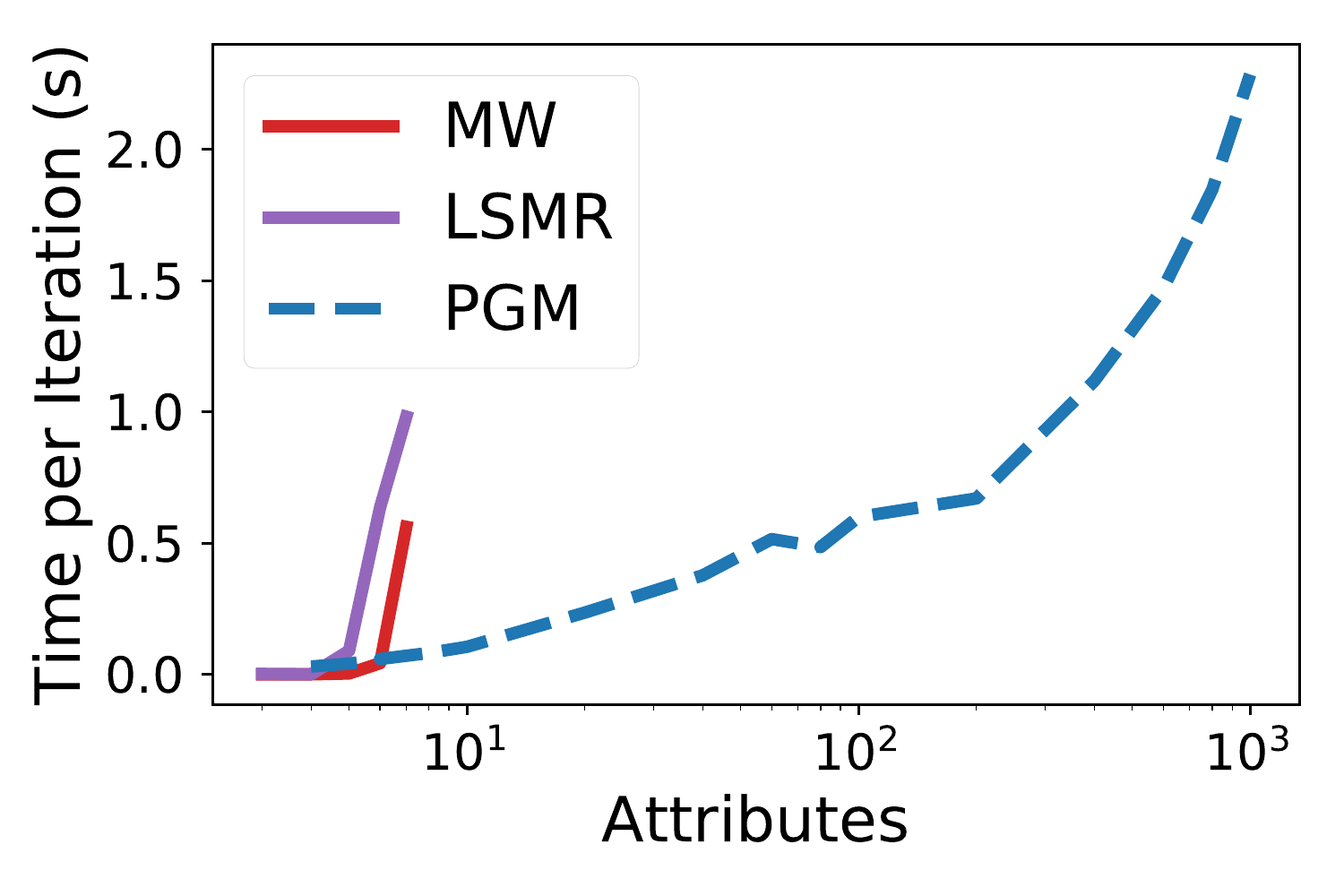} \vspace{-1.5em}
    \caption{\label{fig:scalability}}
\end{subfigure}%
\caption{ (a) Error of HDMM variants on Adult as a function of $\epsilon$. (b) Scalability of estimation algorithms.} 
\end{figure}

\textbf{Varying epsilon.}
While $\epsilon$ is set to 1 in Figure ~\ref{fig:accuracy}, in Figure ~\ref{fig:sensitivity} we look at the impact of varying $\epsilon$, for a fixed dataset and measurement set.  We use the Adult dataset and the measurements selected by HDMM, (which do not depend on $\epsilon$).  The magnitude of the improvement offered by our PGM estimation algorithm increases as $\epsilon$ decreases.  At $\epsilon=0.3$ and below, the mechanism has virtually no utility without PGMs.  At the highest $\epsilon$ of $10.0$, HDMM+LLS actually offers slightly lower error than HDMM+PGM on the workload, although both have very low error in an absolute sense.  The error of HDMM+PGM on the \emph{measurements} is still better by more than a factor of three at this privacy level.  This behavior has been observed before in the low-dimensional setting, where the ordinary least squares estimator generalizes better than the non-negative least squares estimator for workloads with range queries \cite{li2015matrix}.  


\subsection{The scalability of PGM estimation}

We now evaluate the scalability of our approach compared with two other general-purpose estimation techniques: multiplicative weights (MW; \citeauthor{hardt2012simple}, \citeyear{hardt2012simple}) and iterative ordinary least squares (LSMR; \citeauthor{fong2011lsmr}, \citeyear{fong2011lsmr}, \citeauthor{Zhang18Ektelo:}, \citeyear{Zhang18Ektelo:}).  We omit from comparison PrivBayes estimation and DualQuery estimation because they are special-purpose estimation methods that cannot handle arbitrary linear measurements.
We use synthetic data so that we can systematically vary the domain size and the number of attributes.  We measure the marginals for each triple of adjacent attributes --- i.e., $\Q_{C} = \II$ for all $C = (i, i+1, i+2) $ where $1 \leq i \leq d-2 $.  In Figure ~\ref{fig:scalability}, we vary the number of attributes from $ 3 $ to $ 1000 $ (fixing the domain of each attribute, $| \X_i |$ at 10), and plot the time per iteration of each of these estimation algorithms.  Both MW and LSMR fail to scale beyond datasets with $10$ attributes, as they both require materializing $\p$ in vector form, while PGM easily scales to datasets with $1000$ attributes. 

The domain size is the primary factor that determines scalability of the baseline methods.  However, the scalability of PGM primarily depends on the complexity of the measurements taken.  In the experiment above, the measurements were chosen to highlight a case where PGM estimation scales very well.  In general, when the graphical model implied by the measurements has high tree-width, our methods will have trouble scaling, as \oracle{} is computationally expensive.  In these situations, \oracle{} may be replaced with an approximate marginal inference algorithm, like loopy belief propagation \cite{wainwright2008graphical}.

\section{Related Work} \label{sec:related}

The release of linear query answers has been extensively studied by the privacy community~\cite{zhang2017privbayes,li2015matrix,Zhang14Towards,li2014data,Gaboardi14Dual,Yaroslavtsev13Accurate,qardaji2013understanding,Nikolov13Geometry,Thaler12Faster,hardt2012simple,Cormode12Spatial,Acs12Differentially,Gupta11Privately,ding2011differentially,Xiao10Differential,li2010optimizing,hay2010boosting,Hardt10Geometry,Barak07Privacy,mckenna2018optimizing,eugenio2018cipher}.  Early work using inference includes~\cite{Barak07Privacy,hay2010boosting,williams2010probabilistic}, motivated by consistency as well as potential accuracy improvements. Inference has since been widely used in techniques for answering linear queries \cite{lee2015maximum}.  These mechanisms often contain custom specialized inference algorithms that exploit properties of the measurements taken, and can be replaced by our algorithms. 

\cite{williams2010probabilistic} introduce the problem of finding posterior distributions over model parameters from the output of differentially private algorithms. Their problem formulation requires a known model parameterization and a prior distribution over the parameter space.  Their approach requires approximating a high-dimensional integral, which they do either by Markov chain Monte Carlo, or by upper and lower bounds via the ``factored exponential mechanism''.  In the discrete data case, these bounds require summing over the data domain, which is just as hard as materializing $\p$ and is not feasible for high-dimensional data.


\cite{bernstein2017differentially} consider the task of privately learning the parameters of an undirected graphical model. They do so by releasing noisy sufficient statistics using the Laplace mechanism, and then using an expectation maximization algorithm to learn model parameters from the noisy sufficient statistics. Their work shares some technical similarities with ours, but the aims are different. They have the explicit goal of learning a graphical model whose structure is specified in advance and used to determine the measurements. Our goal is to find a compact representation of some data distribution that minimizes a loss function where the measurements are determined externally; the graphical model structure is a by-product of the measurements made and the maximum entropy criterion.

\cite{chen2015differentially} consider the task of privately releasing synthetic data.  Their mechanism is similar to PrivBayes, but it uses undirected graphical models instead of Bayesian networks.  It finds a good model structure using a mutual information criteria, then measures the sufficient statistics of the model (which are marginals) and post-processes them to resolve inconsistencies.  This post-processing is based on a technique developed by \cite{qardaji2014priview} that ensures all measured marginals are internally consistent, and may be improved with our methods.

\clearpage
\bibliography{refs}
\bibliographystyle{icml2019}

\newpage
\appendix
\section{Estimation}

Define $\bmu(\btheta)$ to be the marginals of the graphical model with parameters $\btheta$, which may be computed with the \oracle{}.

\subsection{Proximal Algorithm Derivation}

Our goal is to solve the following optimization problem:
$$ \hat{\bmu} = \argmin_{\bmu \in \marg} L(\bmu) $$
where $L$ is some convex function such as $ \norm{ \Q_\C \bmu - \y } $.  

Using the mirror descent algorithm \cite{beck2003mirror}, we can use the following update equation:
$$\bmu^{t+1} = \argmin_{\bmu \in \marg} \bmu^T \grad L(\bmu^t) + \frac{1}{\eta_t} D(\bmu, \bmu^t)$$
Where $D$ is a Bregman distance measure defined as 
$$D(\bmu, \bmu^t) = \psi(\bmu) - \psi(\bmu^t) - (\bmu - \bmu^t)^T \grad \psi(\bmu^t)$$
for some strongly convex and continuously differentiable function $\psi$.  Using $ \psi = -H$ to be the negative entropy, we arrive at the following update equation:
\begin{align*}
\bmu^{t+1} &= \argmin_{\bmu \in \marg} \bmu^T \grad L(\bmu^t) + \frac{1}{\eta_t} D(\bmu, \bmu^t) \\
&= \argmin_{\bmu \in \marg} \bmu^T \grad L(\bmu^t) + \frac{1}{\eta_t} \Big( -H(\bmu) + \bmu^T \grad H(\bmu^t) \Big) \\
&= \argmin_{\bmu \in \marg} \bmu^T \Big(\grad L(\bmu^t) + \frac{1}{\eta_t} \grad H(\bmu^t) \Big) - \frac{1}{\eta_t} H(\bmu) \\
&= \argmin_{\bmu \in \marg} \bmu^T \Big(\eta_t \grad L(\bmu^t) + \grad H(\bmu^t) \Big) - H(\bmu) \\
&= \argmin_{\bmu \in \marg} \bmu^T \Big(\eta_t \grad L(\bmu^t) - \btheta^t \Big) - H(\bmu) \\
&= \bmu \big( \btheta^t -\eta_t \grad L(\bmu^t) \big)
\end{align*}
The first four steps are simple algebraic manipulation of the mirror descent update equation.  The final two steps use the observation that $ \grad H(\bmu^t) = -\btheta^t $ and that marginal inference can be cast as the following optimization problem: \cite{wainwright2008graphical,vilnis2015bethe}
$$ \bmu(\btheta) = \argmin_{\bmu \in \marg} -\bmu^T \btheta - H(\bmu) $$

Thus, optimization over the marginal polytope is reduced to computing the marginals of a graphical model with parameters $ \btheta^t -\eta_t \grad L(\bmu^t) $, which can be accomplished using belief propagation or some other \oracle.

\subsection{Accelerated Proximal Algorithm Derivation}

The derivation of the accelerated proximal algorithm is similar.  It is based on Algorithm 3 from \cite{xiao2010dual}.  Applied to our setting, step 4 of that algorithm requires solving the following problem:
\begin{align*}
\vect{\nu}^t &= \argmin_{\bmu \in \marg} \bmu^T \vect{\bar{g}} - \frac{4 K}{t (t+1)} H(\bmu) \\
&= \argmin_{\bmu \in \marg} \frac{t(t+1)}{4K} \bmu^T \vect{\bar{g}} - H(\bmu) \\
&= \bmu \Big(-\frac{t (t+1)}{4 L} \bar{\vect{g}} \Big) \\
\end{align*}
which we solve by using the \oracle{}.

\subsection{Direct Optimization}

In preliminary experiments we also evaluated a direct method to solve the optimization problem.  For the direct method, we estimate the parameters $\hat{\btheta}$ directly by reformulating the optimization problem and instead solving the unconstrained problem $ \hat{\btheta} = \argmin_{\btheta} L\big( \bmu(\btheta) \big)$.
To evaluate the optimization objective, we use \oracle{} to compute $\bmu(\btheta)$ and then compute the loss. For optimization, it has been observed that it is possible to backpropagate through marginal inference procedures (with or without automatic differentiation software) to compute their gradients~\cite{eaton2009choosing,domke2013learning}. We apply automatic differentiation to the entire forward computation~\cite{maclaurin2015autograd}, which includes \oracle{}, to compute the gradient of $L$. 


Since this is now an unconstrained optimization problem and we can compute the gradient of $L$, many optimization methods apply. In our experiments, we use $L_2$ loss, which is smooth, and apply the L-BFGS algorithm for optimization~\cite{byrd1995limited}. 

However, despite its simplicity, there is a significant drawback to the direct algorithm. It is not, in general, convex with respect to $\btheta$. This may seem surprising since the original problem is convex, i.e., $L(\bmu)$ is convex with respect to $\bmu$ and $\marg$ is convex. Also, the most well known problem of this form, maximum-likelihood estimation in graphical models, \emph{is} convex with respect to $\btheta$~\cite{wainwright2008graphical}; however, this relies on properties of exponential families that do not apply to other loss functions. One can verify for losses as simple as $L_2$ that the Hessian need not be positive definite. As a result, the direct algorithm is not guaranteed to converge to a global minimum of the original convex optimization problem $\min_{\bmu \in \marg} L(\bmu)$. We did not observe convergence problems in our experiments, but it was not better in practice than the proximal algorithms, which is why it is not included in the paper.


\section{Inference} \label{sec:inference}

\begin{algorithm}[t]
    \caption{Inference for Factored Queries} \label{alg:inference}
\begin{algorithmic}
    \STATE {\bfseries Input:} Parameters $\btheta$, factored query matrix $\Q$
    \STATE {\bfseries Output:} Query answers $\Q\,\p_{\btheta}$
    \STATE $\psi = \{ \exp(\btheta_C) \mid C \in \C \} \cup \{ \Q_i \mid i \in [d] \} $
    \STATE $Z = \oracle(\btheta$)
    \STATE {\bfseries return} $\text{VARIABLE-ELIM}(\psi, \X) / Z$
\end{algorithmic}
\end{algorithm}

We now discuss how to exploit our compact factored representation of $\p_{\btheta}$ to answer new linear queries.
We give an efficient algorithm for answering \emph{factored linear queries}.
\begin{definition}[Factored Query Matrix] \label{def:kron}
  A factored query matrix $\Q$ has columns that are indexed by $\x$ and rows that are indexed by vectors $ \z \in [r_1] \times \dots \times [r_d]$. The total number of rows (queries) is $r = \prod_{i=1}^d r_i$. The entries of $\Q$ are given by
  $\Q(\z, \x) = \prod_{i=1}^d \Q_i(\z_i, \x_i)$,
  where $\Q_i \in \R^{r_i \times n_i}$ is a specified factor for the $i$th attribute. The matrix $\Q$ can be expressed as $\Q = \Q_1 \otimes \dots \otimes \Q_d $, where $\otimes$ is the Kronecker product.
\end{definition}

Factored query matrices are expressive enough to encode any conjunctive query (or a cartesian product of such queries), and more. There are a number of concrete examples that demonstrate the usefulness of answering queries of this form, including:

\begin{itemize}[leftmargin=*,topsep=0pt,itemsep=2pt,partopsep=1pt,parsep=1pt]
\item Computing the marginal $\bmu_C$ for any $C \subseteq [d]$ (including unmeasured marginals).
\item Computing the multivariate CDF of $\bmu_C$ for any $C \subseteq [d]$.
\item Answering range queries.
\item Compressing the distribution by transforming the domain.
\item Computing the (unnormalized) expected value of one variable conditioned on other variables.
\end{itemize}

For the first two examples, we could have used standard variable elimination to eliminate all variables except those in $C$. Existing algorithms are not able to handle the other examples without materializing $\hat{\p}$ (or a marginal that supports the queries). Thus, our algorithm generalizes variable elimination. A more comprehensive set of examples, and details on how to construct these query matrices are given in section \ref{sec:factoredqueries}

The procedure for answering these queries is given in Algorithm~\ref{alg:inference}, which can be understood as follows. For a particular $\z$, write $f(\z, \x) = \Q(\z, \x)\p_{\btheta}(\x) = \prod_i \Q_i(\z_i, \x_i) \p_{\btheta}(\x)$. This can be viewed as an augmented graphical model on the variables $\z$ and $\x$ where we have introduced new pairwise factors between each $(\x_i, \z_i)$ pair defined by the query matrix. Unlike a regular graphical model, the new factors can contain negative values. The query answers are obtained by multiplying $\Q$ and $\p$, which sums over $\x$. The $\z$th answer is given by:
\begin{align*}
(\Q \p_{\btheta})(\z) &= \sum_{\x \in \X} \Q(\z,\x) \p_{\btheta}(\x) \\
&= \frac{1}{Z} \sum_{\x \in \X} \prod_{i=1}^d \Q_i(\z_i, \x_i) \prod_{C \in \C} \exp[\btheta_C(\x_C)]
\end{align*}
This can be understood as marginalizing over the $\x$ variables in the augmented model $f(\z, \x)$. The {\sc Variable-Elim} routine referenced in the algorithm is standard variable elimination to perform this marginalization; it can handle negative values with no modification. We stress that, in practice, factor matrices $\Q_i$ may have only one row ($r_i = 1$, e.g., for marginalization); hence the output size $r = \prod_{i=1}^d r_i$ is not necessarily exponential in $d$.

\eat{
As shown in algorithm~\ref{alg:inference}, we use variable elimination over an augmented model with factors from the original model combined with the $\Q_i$ matrices interpreted as two variable factors.  The $\Q_i$ factors introduce new variables $\z_1, \dots, \z_d$.  The correctness of the algorithm can be seen by analyzing the following sum:

$$(\Q \p_{\btheta})(\z) = \sum_{\x \in \X} \Q(\z,\x) \p_{\btheta}(\x) = \frac{1}{Z} \sum_{\x \in \X} \prod_{i=1}^d \Q_i(\z_i, \x_i) \prod_{C \in \C} \exp[\btheta_C(\x_C)] $$

Of course, evaluating the sum directly is intractable when $| \X |$ is large, but this is exactly the type of sum that can be evaluated efficiently using variable elimination (when the factorization allows it).  \ry{more details might be necessary here... Dan what are your thoughts?}
}

\subsection{Factored Query Matrices} \label{sec:factoredqueries}

\begin{table*}[t]
\centering
\begin{tabular}{|l|c|c|l|l|}
$\Q_i$ & Requirements & Size & Definition $(\forall a \in [n_i])$ & Description \\\hline
$\II$ & & $n_i \times n_i $ & $\Q_i(a,a) = 1$ & keep variable in \\
$\T$ & & $1 \times n_i$ & $\Q_i(1,a) = 1$ & marginalize variable out \\
$\vect{e}_j$ & $j \in [n_i]$ & $1 \times n_i$ & $\Q_i(1,j) = 1$ & inject evidence \\
$\vect{e}_S$ & $S \subseteq [n_i]$ & $1 \times n_i$ & $\Q_i(1,j) = 1 \:\forall j \in S$ & inject evidence (disjuncts)\\
$\P$ & & $n_i \times n_i$ & $\Q_i(b,a) = 1 \:\forall b \geq a$ & transform into CDF \\
$\RR_f$ & $f : [n_i] \rightarrow [r_i]$ & $r_i \times n_i$ & $\Q_i(f(a), a) = 1$ & compress domain \\
$\mathbf{E}$ & & $1 \times n_i$ & $\Q_i(1, a) = a$ & reduce to mean \\
$\mathbf{E}_k$ & $k \geq 1$ & $ k \times n_i$ & $\Q_i(b,a) = a^b \: \forall b \leq k$ & reduce to first k moments \\\hline
\end{tabular}
\caption{Example factors in the factored query matrix} \label{table:factored}
\end{table*}

Table~\ref{table:factored} gives some example ``building block'' factors that can be used to construct factored query matrices.  This is by no means an exhaustive list of possible factors but it provides the reader with evidence that answering these types of queries efficiently is practically useful.  The factored query matrix for computing the marginal $\bmu_C$ uses $\Q_{i} = \II $ for $ i \in C $ and $ \Q_i = \T$ for $i \notin C$.  Similarly, the factored query matrix for computing the multivariate CDF of $\bmu_C$ would simply use $\Q_i = \P$ for $i \in C$.  A query matrix for compressing a distribution could be characterized by functions $ f_i : [n_i] \rightarrow [2] $ or equivalently binary matrices $ \Q_i = \RR_{f_i} \in \R^{2 \times n_i} $.   The query matrix for computing the (unnormalized) expected value of variable $i$ conditioned on variable $j$ would use $ \Q_i = \mathbf{E} $ and $ \Q_j = \II$ (and $\Q_k = \T$ for all other $k$).  These are only a few examples; these building blocks can be combined arbitrarily to construct a wide variety of interesting query matrices.  

\section{Loss Functions}

\subsection{$L_1$ and $L_2$ losses}

The $L_1$ and $L_2$ loss functions have simple (sub)gradients.
\begin{align*}
\grad L_1 (\bmu) &= \Q_{\C}^T \text{sign}(\Q_{\C} \bmu - \y) \\
\grad L_2(\bmu) &= \Q_{\C}^T (\Q_{\C} \bmu - \y)
\end{align*}

\subsection{Linear measurements with unequal noise}

When the privacy budget is not distributed evenly to the measurements in the we have to appropriately modify the loss functions, which assume that the noisy answers all have equal variance. 
In order to do proper estimation and inference we have to account for this varying noise level in the loss function.  In section ~\ref{sec:formulation} we claimed that $L(\p) = \norm{ \Q \p - \y }$ makes sense as a loss function when the noise introduced to $\y$ are iid.  Luckily, even if this assumption is not satisfied it is easy to correct.  Assume that $ y_i = \q_i^T \p + \varepsilon_i $ where $ \varepsilon_i \sim Lap(b_i) $.  Then $ \frac{1}{b_i} y_i = \frac{1}{b_i} \q_i^T \p + \frac{1}{b_i} \varepsilon_i $ and $ \frac{1}{b_i} \varepsilon_i \sim Lap(1) $.  Thus, we can replace the query matrix $ \Q \leftarrow \mathbf{D} \Q$ and the answer vector $ \y \leftarrow \mathbf{D} \y $ where $ \mathbf{D} $ is the diagonal matrix defined by $\mathbf{D}_{ii} = \frac{1}{b_i}$.  All the new query answers have the same effective noise scale, and so the standard loss functions may be used.  This idea still applies if the noise on each query answer is sampled from a normal distribution as well (for $(\epsilon, \delta)$-differential privacy).    

\subsection{Dual Query Loss Function}

Algorithm \ref{alg:dq} shows DualQuery applied to workloads defined over the marginals of the data.  There are five hyper-parameters, of which four must be specified and the remaining one can be determined from the others. 

The first step of the algorithm computes the answers to the workload queries.  Then for $T$ time steps observations are made about the true data via samples from the distribution $Q^t$.  These observations are used to find a record $ \x \in \X $ to add to the synthetic database.

\begin{algorithm}[H]
    \caption{Dual Query for marginals workloads} \label{alg:dq}
\begin{algorithmic}
    \STATE {\bfseries Input:} $\vect{X}$, the true data
    \STATE {\bfseries Input:} $\W_{\C}$, workload queries
    \STATE {\bfseries Input:} $(s, T, \eta$, $\epsilon$, $\delta$), hyper-parameters
    \STATE {\bfseries Output:} synthetic database of $T$ records
    \STATE $ \y = \W_C \bmu_{\vect{X}} $
    \STATE $ Q^1 = \text{uniform}(\W) $
    \FOR{$t = 1, \dots, T$}
    \STATE sample $\q^t_1, \dots \q^t_s$ from $Q^t$
    \STATE $ \x^t = \argmax_{\x \in \X} \sum_{i=1}^s \q^t_i \bmu - \q^t_i \bmu_{\x} $
    \STATE $ Q^{t+1} = Q^t \odot \exp{(-\eta*(\y - \W_C \bmu_{\x^t}))} $
    \STATE normalize $Q^t$
    \ENDFOR
    \STATE {\bfseries return} $(\x^1, \dots, \x^T)$
\end{algorithmic}
\end{algorithm}

Algorithm \ref{alg:dqloss} shows a procedure for computing the negative log likelihood (our loss function) of observing the DualQuery output, given some marginals.  Evaluating the log likelihood is fairly expensive, as it requires basically simulating the entire DualQuery algorithm.  Fortunately we do not have to run the most computationally expensive step within the procedure, which is finding $\x^t$.  We differentiate this loss function using automatic differentiation \cite{maclaurin2015autograd} for use within our estimation algorithms.

\begin{algorithm}[H]
    \caption{Dual Query Loss Function} \label{alg:dqloss}
\begin{algorithmic}
    \STATE {\bfseries Input:} $\bmu$, marginals of the data
    \STATE {\bfseries Input:} $\W_{\C}$, workload queries
    \STATE {\bfseries Input:} cache, all relevant output from DualQuery \\
        > $\q^t_1, \dots \q^t_s$ - sampled queries at each time step \\
        > $\x^t$ - chosen record at each time step \\
    \STATE {\bfseries Output:} $L(\bmu)$, the negative log likelihood
    \STATE $ \y = \W_C \bmu $
    \STATE $ Q^1 = \text{uniform}(\W) $
    \STATE loss $= 0$
    \FOR{$t = 1, \dots, T$}
    \STATE  loss --$= \sum_{i=1}^s \log{(Q^t(\q^t_i))} $
    \STATE $ Q^{t+1} = Q^t \odot \exp{(-\eta*(\y - \W_C \bmu_{\x^t}))} $
    \STATE normalize $Q^t$
    \ENDFOR
    \STATE {\bfseries return} loss
\end{algorithmic}
\end{algorithm}

\section{Additional Experiments}

\subsection{$L_1$ vs. $L_2$ Loss}

\begin{figure*}
\centering
\includegraphics[width=0.33\textwidth]{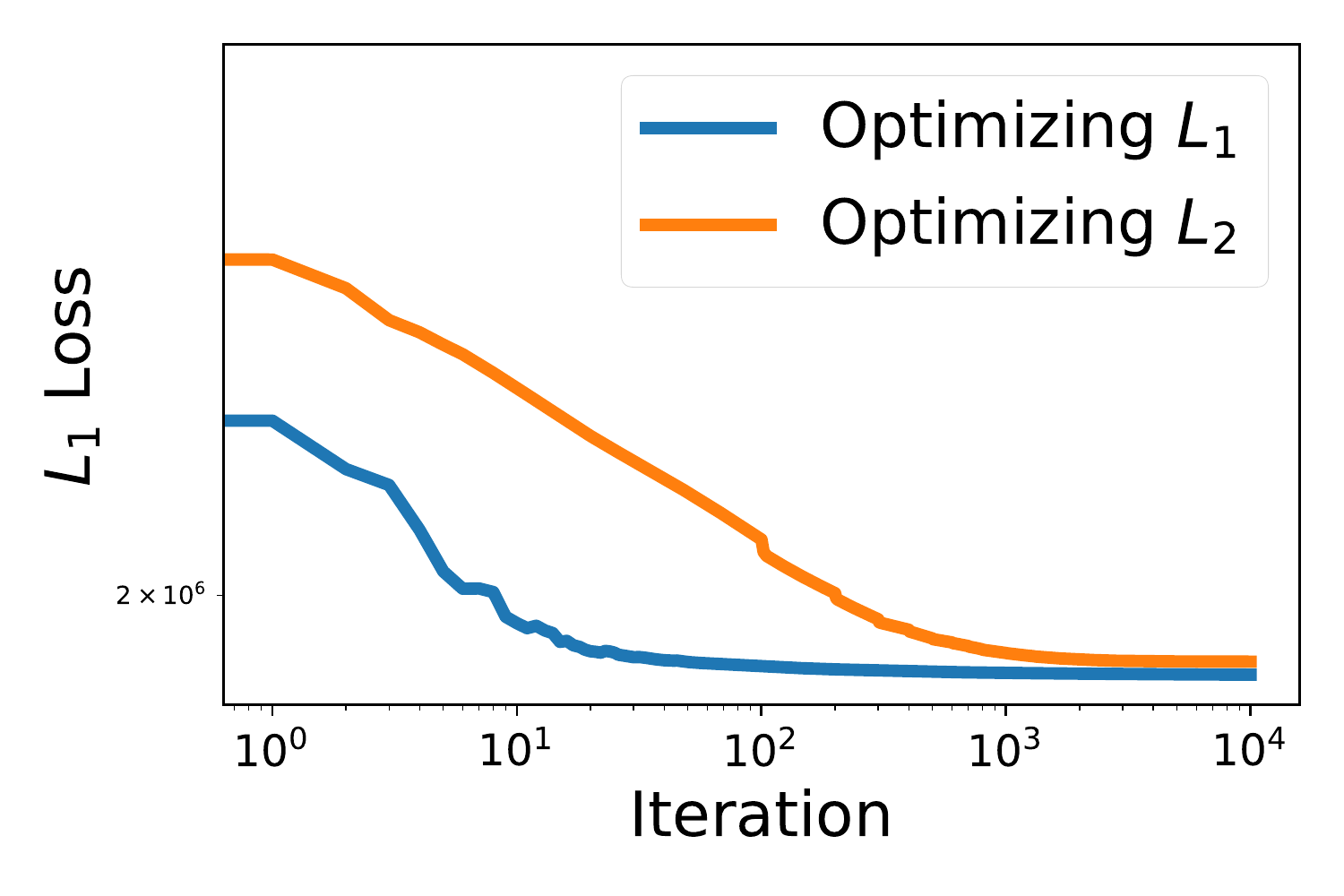}
\includegraphics[width=0.33\textwidth]{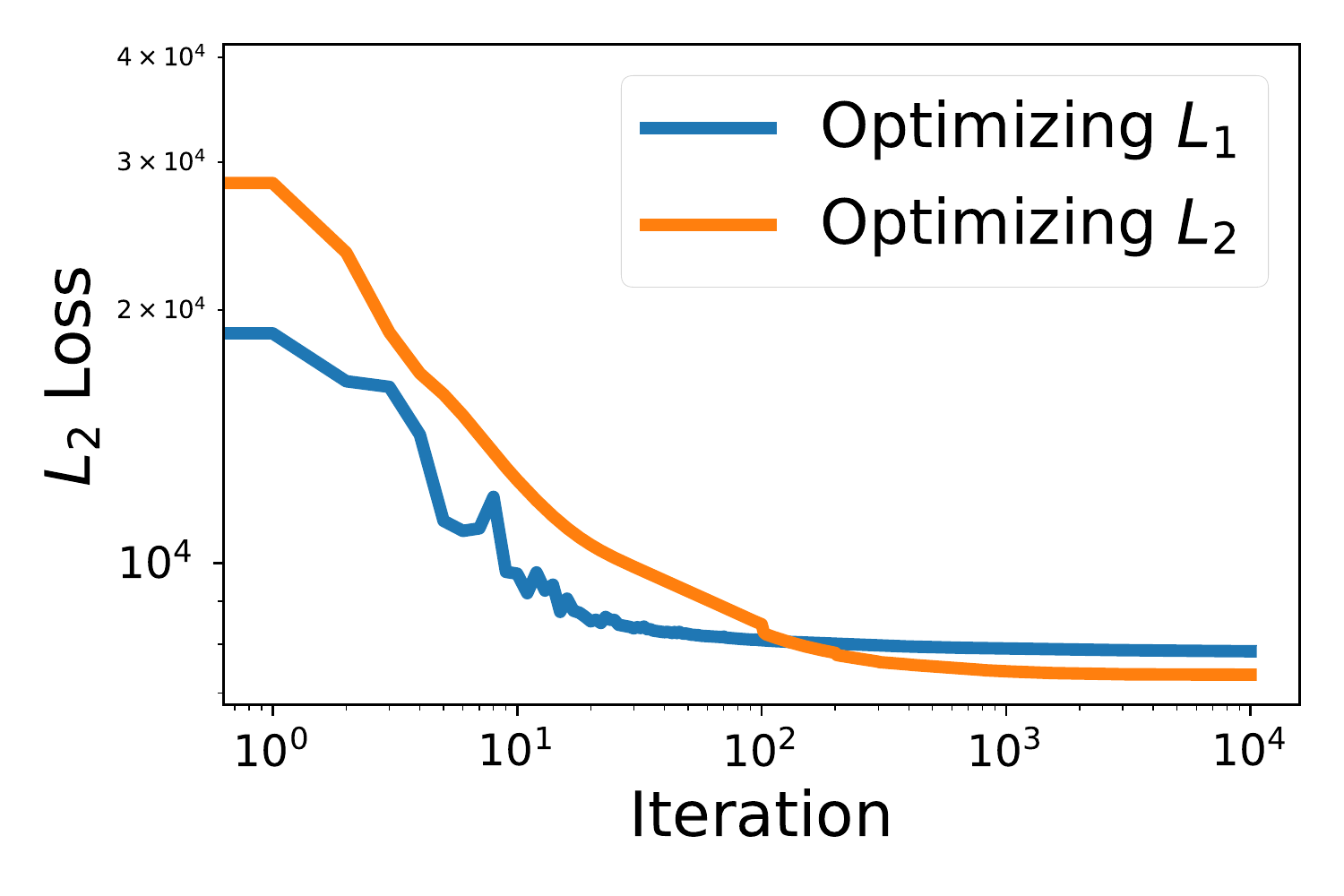}
\includegraphics[width=0.33\textwidth]{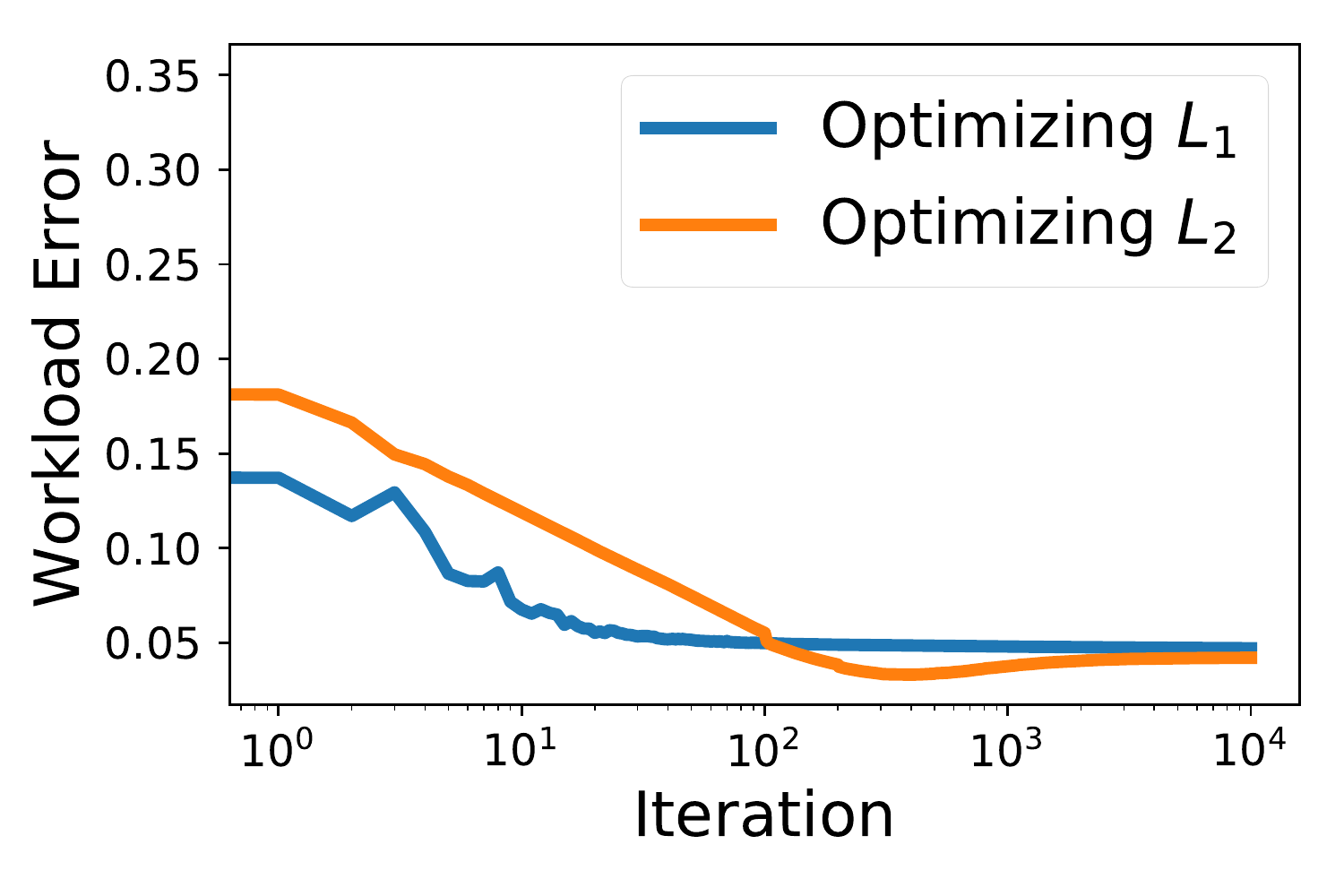}
\caption{$L_1$ minimization vs. $L_2$ minimization, evaluated on $L_1$ loss, $L_2$ loss, and workload error} \label{fig:l1_vs_l2}
\end{figure*}

In Section ~\ref{approach} we mentioned that minimizing $L_1$ loss is equivalent to maximizing likelihood for linear measurements with Laplace noise, but that $L_2$ loss is more commonly used in the literature.  In this experiment we compare these two estimators side-by-side.  Specifically, we consider the workload from Figure ~\ref{fig:accuracy} and measurements chosen by HDMM with $\epsilon = 1.0$.   As expected, performing $L_1$ minimization results in lower $L_1$ loss but higher $L_2$ loss, although the difference is quite small, especially for $L_1$ loss.  The difference is larger for $L_2$ loss.  Minimizing $L_2$ loss results in lower workload error, indicating that it generalizes better.  This is somewhat surprising given that $L_1$ minimization is maximizing likelihood.  Another interesting observation is that the workload error actually starts going up after about 200 iterations, suggesting that some form of over-fitting is occurring.  There minimum workload error achieved was $ 0.066 $ while the final workload error was $0.084$ --- a pretty meaningful difference.  Of course, in practice we cannot stop iterating when workload error starts increasing because evaluating it requires looking at the true data.

\section{Additional Details}

\subsection{Unknown Total}

Our algorithms require $m$, the total number of records in the dataset is known or can be estimated.  Under a slightly different privacy definition where $\nbrs{(\db)}$ is the set of databases where a single record is added or removed (instead of modified), this total is a sensitive quantity which cannot be released exactly \cite{Dwork14Algorithmic}.  Thus, the total is not known in this setting, but a good estimate can typically be obtained from the measurements taken, without spending additional privacy budget.  First observe that $ \vect{1}^T \bmu_C = m$ is the total for an unnormalized database.  Now suppose we have measured $\y_C = \Q_C \bmu_C + \z_C $.  Then as long as $ \vect{1}^T $ is in the row-space of $\Q_C$, $ m_C = \vect{1}^T \Q_C^+ \y_C $ is an unbiased estimate for $m$ with variance $ Var(m_C) = Var(\y_C) \norm{\vect{1}^T \Q_C^+}_2^2 $. This is a direct consequence of Proposition 9 from \cite{li2015matrix}.  We thus have multiple estimates for $m$ which we can combine using inverse variance weighting, resulting in the final estimate of $\hat{m} = \frac{\sum_C m_C / Var(m_C)}{\sum_C 1 / Var(m_C)} $, which we can use in place of $m$.

\subsection{Multiplicative Weights vs Entropic Mirror Descent} \label{sec:mw_emd}

Recall from Section \ref{sec:methods} that the multiplicative weights update equation is:
$$ \hat{\p} \leftarrow \hat{\p} \odot \exp{(\q_i (\q_i^T \hat{\p} - y_i)) / 2m} / Z $$
and the update is applied (possibly cyclically) for $i=1, \dots, T$.  Now imagine taking all of the measurements and organizing them into a $ T \times n$ matrix $\Q$.  Then we can apply all the updates at once, instead of sequentially, and we end up with the following update equation.
$$ \hat{\p} \leftarrow \hat{\p} \odot \exp{(\Q^T (\Q \hat{\p} - \y) / 2m)} / Z $$
Observing that $ \grad L_2(\hat{\p}) = \Q^T (\Q \hat{\p} - \y) $, this simplifies to:
$$ \hat{\p} \leftarrow \hat{\p} \odot \exp{(\grad L_2(\hat{\p}) / 2m)} / Z $$
which is precisely the update equation for entropic mirror descent for minimizing $L_2(\p)$ over the probability simplex \cite{beck2003mirror}.

\end{document}